\documentclass[journal]{IEEEtran}
\usepackage{amssymb}
\usepackage{multirow}
\usepackage{xcolor}
\usepackage[normalem]{ulem}
\usepackage{graphicx}
\usepackage{listings}
\usepackage[most]{tcolorbox}
\usepackage{longtable} 
\usepackage{array} 
\usepackage{tabularx}
\usepackage{booktabs}
\usepackage{caption}
\usepackage{amsmath}
\usepackage{booktabs}
\usepackage[bottom=0.98in,top=0.75in,left=0.625in,right=0.625in]{geometry}

\begin{document}

\title{Privacy-Preserving and Scalable IoV Management System Using Open-Source LLMs}
\title{FPoTT: An Open-Source LLM-Driven Federated Transformer for Predictive IoV Management}

\title{Open-Source LLM-Driven Federated Transformer for Predictive IoV Management}

\author{
    \IEEEauthorblockN{
    Yazan Otoum\IEEEauthorrefmark{1}, Arghavan Asad\IEEEauthorrefmark{1}, Ishtiaq Ahmad\IEEEauthorrefmark{4} \\} \vspace{1em}
  \IEEEauthorblockA{\small \IEEEauthorrefmark{1}School of Computer Science and Technology, Algoma University, Canada}
        \\ \IEEEauthorblockA{\small \IEEEauthorrefmark{4}Faculty of Electrical Engineering, Czech Technical University in Prague, Czech Republic}
    
        }

\maketitle

\thispagestyle{empty}
\pagestyle{empty}

\begin{abstract}


The proliferation of connected vehicles within the Internet of Vehicles (IoV) ecosystem presents critical challenges in ensuring scalable, real-time, and privacy-preserving traffic management. Existing centralized IoV solutions often suffer from high latency, limited scalability, and reliance on proprietary Artificial Intelligence (AI) models, creating significant barriers to widespread deployment, particularly in dynamic and privacy-sensitive environments. Meanwhile, integrating Large Language Models (LLMs) in vehicular systems remains underexplored, especially concerning prompt optimization and effective utilization in federated contexts. To address these challenges, we propose the Federated Prompt-Optimized Traffic Transformer (FPoTT), a novel framework that leverages open-source LLMs for predictive IoV management. FPoTT introduces a dynamic prompt optimization mechanism that iteratively refines textual prompts to enhance trajectory prediction. The architecture employs a dual-layer federated learning paradigm, combining lightweight edge models for real-time inference with cloud-based LLMs to retain global intelligence. A Transformer-driven synthetic data generator is incorporated to augment training with diverse, high-fidelity traffic scenarios in the Next Generation Simulation (NGSIM) format. Extensive evaluations demonstrate that FPoTT, utilizing EleutherAI Pythia-1B, achieves 99.86\% prediction accuracy on real-world data while maintaining high performance on synthetic datasets. These results underscore the potential of open-source LLMs in enabling secure, adaptive, and scalable IoV management, offering a promising alternative to proprietary solutions in smart mobility ecosystems.
\end{abstract}


\section{Introduction}

The rapid evolution of the Internet of Vehicles (IoV) is revolutionizing modern transportation systems by enabling real-time vehicular communication, autonomous decision-making, and intelligent traffic coordination. As the number of connected vehicles surges, the volume of data generated grows exponentially, necessitating efficient data management and real-time analytics to maintain optimal performance and security. However, existing IoV management solutions face significant scalability, interoperability, and data privacy challenges, hindering their widespread deployment in real-world scenarios \cite{gerla2014internet}. Traditional centralized cloud-based approaches introduce substantial latency, rendering them unsuitable for real-time decision making in dynamic traffic conditions. Moreover, proprietary Artificial Intelligence (AI) driven IoV systems often rely on closed-source Large Language Models (LLMs), raising concerns regarding privacy, security, and cost constraints. These limitations underscore the need for an alternative approach that ensures scalable, privacy-preserving, and cost-effective IoV management \cite{haddaji2024iov}.
Recent advancements have explored various methodologies to address these 
challenges, including the need for scalable system architectures, low-latency decision-making, and enhanced privacy protection in data-intensive and distributed IoV environments. For instance, the authors of~\cite{chen2025mobility} proposed a mobility-aware decentralized, federated learning framework that integrates joint resource allocation to enhance vehicular network performance. Their approach emphasizes the importance of decentralization in reducing latency and improving scalability within IoV systems. Similarly, the authors of~\cite{manh2024homomorphic} introduced a homomorphic encryption-enabled federated learning framework to preserve privacy in intrusion detection systems for resource-constrained IoV networks. This method highlights the critical role of encryption techniques in safeguarding vehicular data. 
Building on these efforts, recent attention has turned to the use of LLMs in IoV systems, particularly those that are open-source and adaptable. In this context, Meta's release of Llama 4 \cite{meta2025llama4} represents a significant milestone. Llama 4 is a multimodal AI system capable of processing and integrating various data types, including text, video, images, and audio. Its open-source nature offers transparency and adaptability, making it a viable candidate for secure and decentralized IoV applications. Additionally, DeepSeek's release of DeepSeek-V2 \cite{deepseek2024v2} demonstrates that high-performing language models can be developed efficiently in an open-source manner, offering competitive capabilities while remaining accessible for integration into edge and federated environments. Despite recent advances, open-source LLM integration in real-time, privacy-sensitive IoV environments remains limited. Current solutions lack effective prompt optimization, struggle with low-latency inference on constrained devices, and underutilize synthetic data for generalization. To address these gaps, we introduce Federated Prompt-Optimized Traffic Transformer (FPoTT), a framework that unifies open-source LLMs, dynamic prompt tuning, federated learning, and Transformer-based synthetic data generation. FPoTT enhances trajectory prediction through feedback-driven prompt refinement, achieves scalable, privacy-preserving inference via a dual-layer cloud-edge architecture, and improves generalization using high-fidelity NGSIM-style synthetic scenarios. Evaluations with EleutherAI Pythia-1B show 99.86\% accuracy on real-world data, confirming FPoTT’s effectiveness for secure and adaptive IoV management. The key contributions of this work include:
\begin{itemize}
    \item A novel IoV management framework, FPoTT, that integrates open-source LLMs with dynamic prompt optimization to enhance predictive trajectory modelling for connected vehicles.

    \item A dual-layer federated learning architecture that enables real-time, privacy-preserving inference at the edge while supporting scalable global model aggregation at the cloud.

    \item A Transformer-based synthetic traffic generator that augments real-world datasets with diverse, high-fidelity scenarios to improve model generalization in dynamic IoV environments.

    \item A privacy-first federated training strategy using localized learning, model distillation, and parameter quantization to ensure compliance with data protection regulations and support resource-constrained deployments.

    \item A comprehensive experimental evaluation demonstrating state-of-the-art trajectory prediction performance, achieving up to 99.86\% accuracy on real-world and synthetic datasets with lightweight, open-source LLMs (e.g., Pythia-1B).
\end{itemize}

The remainder of this paper is organized as follows: Section~\ref{sec:literature_review} reviews relevant prior works.  Section~\ref{sec:proposed_framework} describes the architecture and implementation details of the proposed FPoTT framework, including cloud-edge integration and open-source LLM optimizations. Section~\ref{sec:results_analysis} presents experimental evaluations and simulation results to demonstrate the effectiveness of our approach. Finally, Section~\ref{sec:conclusion} concludes the paper and outlines directions for future research.

\begin{figure}[htbp]
    \centering
    \includegraphics[width=1.0\columnwidth]{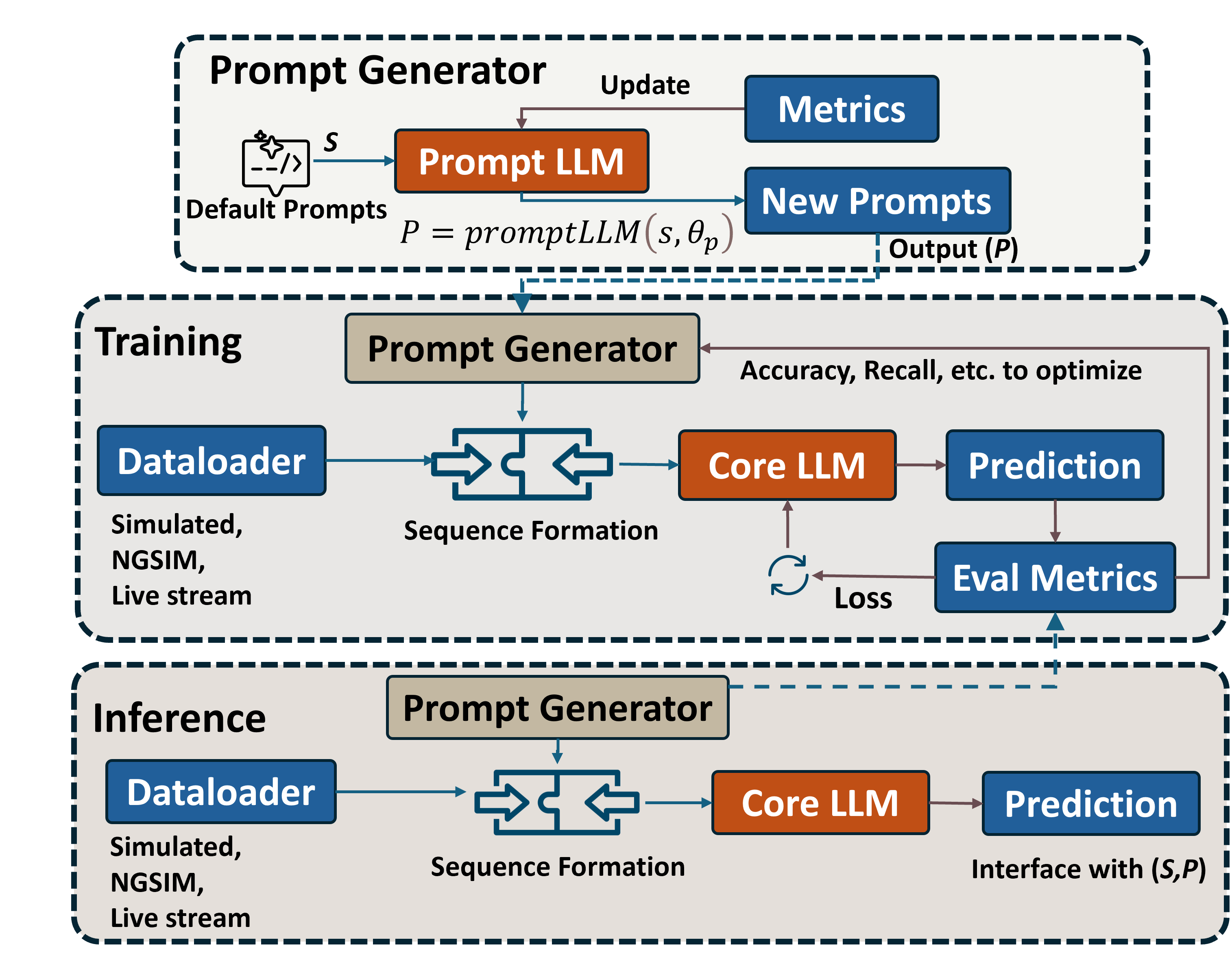}
    \caption{Simplified Overview of the proposed framework. The system consists of three components: a prompt optimizer that refines input prompts based on evaluation metrics, a training pipeline using both real and synthetic NGSIM data to fine-tune LLMs, and an inference stage for real-time traffic prediction.}
    \label{fig:arch}
\end{figure}
\vspace{-0.5cm}

\vspace{-0.5cm}
\section{Related Works} 
\label{sec:literature_review}
Recent advancements in Internet of Things (IoT) environments have increasingly leveraged AI techniques to optimize network performance, enable intelligent decision-making, and enhance privacy compliance \cite{otoum2025llms, otoum2024advancing}. As these systems generate large volumes of sensitive data across distributed nodes, the need for decentralized and privacy-aware solutions has grown. In response, the IoV domain has embraced federated learning and encryption-based frameworks, which offer effective mechanisms for secure, collaborative model training without exposing raw data. 
The paper~\cite{xing2025federated} introduced an adaptive federated learning method that dynamically adjusts learning rates based on real-time vehicular conditions. The work~\cite{qian2024toward} proposed a sparse federated learning technique for object detection, significantly reducing communication overhead while maintaining high detection accuracy. More recently, the authors of~\cite{alqubaysi2025federated} presented a predictive traffic management framework featuring a privacy-preserving scheme to protect sensitive vehicle data against adversarial threats. The work~\cite{xu2024federated} explored federated learning with Unmanned Aerial Vehicle (UAV) swarms to enhance vehicular data collection and processing efficiency and scalability. The integration of LLMs and advanced simulators has further expanded the capabilities of IoV management systems. The paper~\cite{sharma2024openroad} demonstrated the application of LLaMA models for vehicular data processing, emphasizing their efficiency in resource-constrained environments. On the simulation front, the authors of~\cite{protzmann2022implementation} utilized Eclipse MOSAIC to validate cloud-edge IoV architectures. Complementing these simulation-based approaches, the authors of~\cite{you2024real} highlight the importance of grounding simulated traffic environments in real-world data. By integrating real traffic datasets into their simulation workflows, they enhanced the fidelity of synthetic scenarios and improved the generalizability of AI models trained on them. This line of research illustrates a growing trend in IoV systems: bridging the gap between simulated environments and real-world conditions to enable more robust and adaptive models. These developments collectively signal a paradigm shift in AI-driven IoV management from isolated, privacy-focused solutions to integrated frameworks that combine real-world awareness, scalable learning, and LLM capabilities. FPoTT unifies these advancements into a deployable, open-source system optimized for scalable, privacy-preserving IoV management. The following section outlines the framework’s components and how they collectively address current limitations in trajectory prediction, model adaptability, and system decentralization.

\section{Proposed Framework FPoTT}
\label{sec:proposed_framework}
The rapid expansion of connected and autonomous vehicles within the IoV ecosystem necessitates real-time, privacy-aware, and scalable traffic management solutions. While prior research has explored federated learning and simulation environments for vehicular networks, existing methods often suffer from three critical limitations: reliance on proprietary or closed-source models, lack of task-specific optimization strategies for LLMs, and insufficient diversity in training datasets due to real-world data scarcity. To address these challenges, FPoTT is designed as an open-source framework that synergizes prompt optimization, federated learning, and synthetic data generation. FPoTT enhances trajectory prediction and decision-making in real-time IoV environments while maintaining data privacy and model adaptability. This section presents the architectural and functional design of FPoTT, which integrates a feedback-driven prompt optimization mechanism, a dual-layer federated learning framework for cloud-edge collaboration, a Transformer-based synthetic data generator to enhance generalization, and a unified training and inference pipeline for real-time, privacy-preserving trajectory prediction. An overview of the proposed framework architecture is provided in Fig.~\ref{fig:arch}.

\subsection{Prompt Optimization Mechanism for Trajectory Prediction}


Effective trajectory prediction in IoV systems requires accurate models and well-crafted inputs. In LLM-based architectures, transforming traffic data into textual prompts is critical for prediction performance. However, existing approaches typically rely on static or manually designed prompts, which lack adaptability to dynamic traffic conditions. To address this limitation, FPoTT introduces a novel prompt optimization mechanism, an integral part of our contribution, that automatically refines prompts using a dedicated Prompt Generator LLM. This module takes sequential traffic data (e.g., position, velocity, acceleration logs from NGSIM or similar real-world sources). It transforms it into natural language prompts optimized for the trajectory prediction task. The prompts are iteratively refined by evaluating prompt effectiveness using metrics such as accuracy and recall during model training. Mathematically, the prompt generation process is defined as follows: Given a sequence of traffic data points \(\mathcal{S} = \{s_1, s_2, \dots, s_n\}\), the Prompt Generator LLM produces a prompt \(P\) represented by:
\begin{equation}
    P = \text{PromptLLM}(\mathcal{S}, \theta_p),
\end{equation}
where \(\theta_p\) denotes the parameters of the Prompt Generator LLM. During training, evaluation metrics \(\mathcal{M}\), such as accuracy and recall, are utilized to update the Prompt Generator parameters iteratively:
\begin{equation}
    \theta_p \leftarrow \theta_p - \eta \nabla_{\theta_p}\mathcal{L}(\mathcal{M}),
\end{equation}
where \(\eta\) is the learning rate, and \(\mathcal{L}\) denotes the optimization loss function based on metrics. Unlike conventional prompt tuning, which typically uses static templates or fine-tunes prompt embeddings, our Prompt Generator iteratively evaluates and refines prompts based on real-time feedback from evaluation metrics. We conduct a maximum of 50 refinement iterations per batch, with prompts selected based on their ability to maximize validation accuracy and minimize loss. This dynamic, feedback-driven optimization ensures context-specific tailoring of prompts aligned with the target task. 


\begin{figure}[htp!]
    \centering
    \includegraphics[scale=0.50]{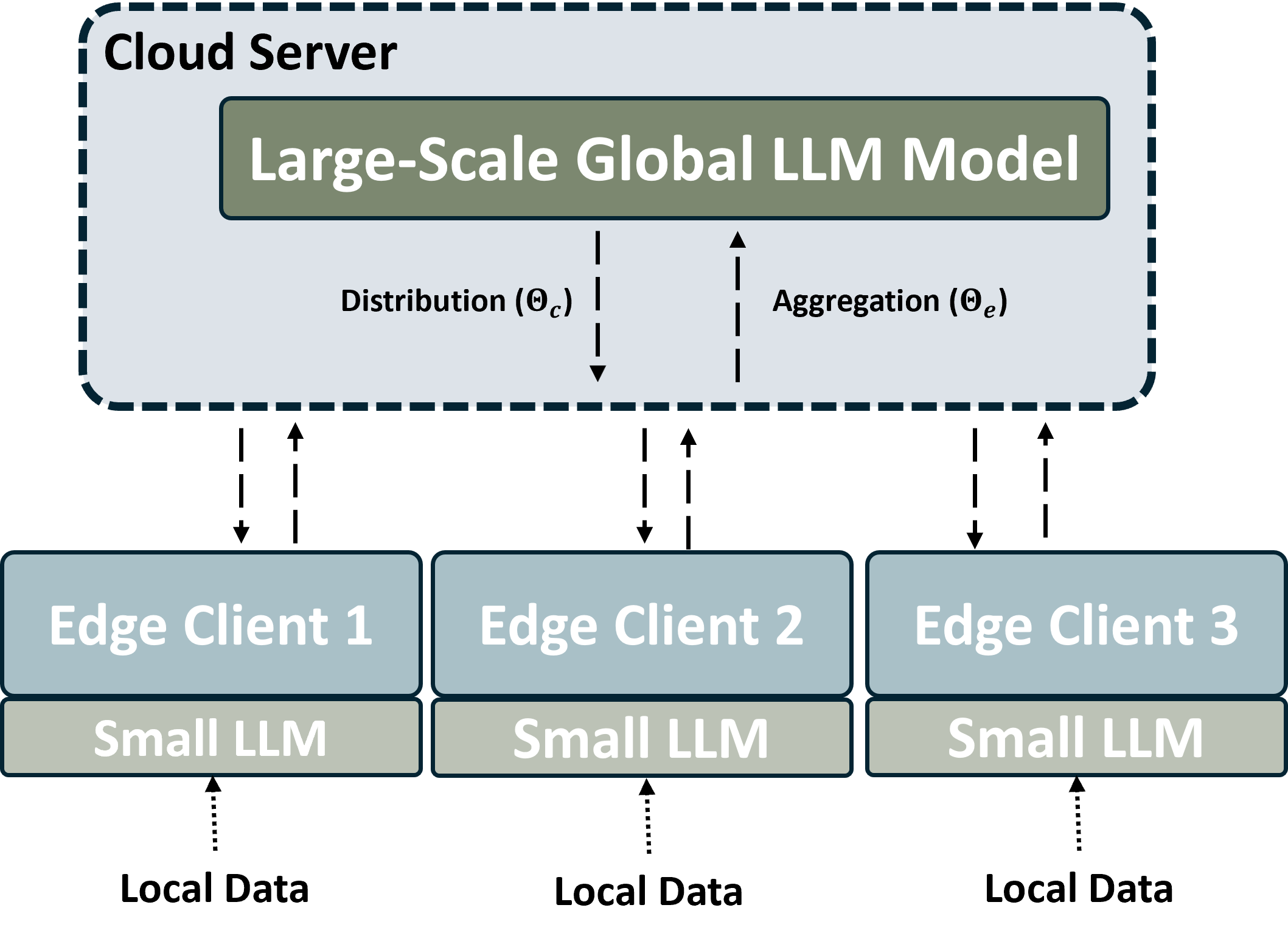} \caption{Illustration of FPoTT's federated learning architecture where a central cloud server distributes and aggregates model parameters $\theta_c$, $\theta_e$ to and from edge clients running local small LLMs on private local data. In real-world applications, several clients could be running under the same server.}
    \label{fig:fed}
\vspace{-0.8cm}
\end{figure}

\begin{table*}[ht]
\centering
\caption{Performance Comparison on NGSIM and Simulated Datasets. Each row reports precision, recall, F1-score, and accuracy.}
\label{tab:performance_comparison_combined}
\begin{tabular}{l|cccc|cccc}
\toprule
\multirow{2}{*}{\textbf{Models}} & \multicolumn{4}{c|}{\textbf{NGSIM}} & \multicolumn{4}{c}{\textbf{Simulated Dataset}} \\
 & \textbf{Precision} & \textbf{Recall} & \textbf{F1} & \textbf{Accuracy} 
 & \textbf{Precision} & \textbf{Recall} & \textbf{F1} & \textbf{Accuracy} \\
\midrule
BERT                   & 0.9735 & 0.9711 & 0.9672 & 0.9705 & 0.9387 & 0.9315 & 0.9280 & 0.9302 \\
ALBERT                 & 0.9762 & 0.9724 & 0.9690 & 0.9763 & 0.9425 & 0.9352 & 0.9311 & 0.9406 \\
PALM                   & 0.9648 & 0.9610 & 0.9605 & 0.9624 & 0.9281 & 0.9210 & 0.9195 & 0.9228 \\
ROBERTA                & 0.9660 & 0.9607 & 0.9580 & 0.9676 & 0.9299 & 0.9207 & 0.9181 & 0.9276 \\
DEBERTA                & 0.9694 & 0.9670 & 0.9645 & 0.9693 & 0.9353 & 0.9280 & 0.9260 & 0.9342 \\
XLNET                  & 0.9719 & 0.9686 & 0.9653 & 0.9724 & 0.9376 & 0.9304 & 0.9269 & 0.9363 \\
GPT-Neo                & 0.9686 & 0.9646 & 0.9604 & 0.9674 & 0.9342 & 0.9265 & 0.9224 & 0.9309 \\
GPT-2 (124M)           & 0.9799 & 0.9771 & 0.9728 & 0.9785 & 0.9410 & 0.9331 & 0.9293 & 0.9380 \\
GPT-2-Medium (355M)    & 0.9732 & 0.9804 & 0.9762 & 0.9838 & 0.9513 & 0.9440 & 0.9402 & 0.9510 \\
GPT-2-Large (774M)     & 0.9807 & 0.9780 & 0.9774 & 0.9845 & 0.9489 & 0.9417 & 0.9389 & 0.9496 \\
flan-t5-base           & 0.7996 & 0.7018 & 0.6842 & 0.7885 & 0.7653 & 0.6710 & 0.6528 & 0.7542 \\
flan-t5-large          & 0.8149 & 0.7056 & 0.6926 & 0.7917 & 0.7784 & 0.6799 & 0.6630 & 0.7596 \\
pythia-410M            & 0.9799 & 0.9764 & 0.9732 & 0.9833 & 0.9456 & 0.9380 & 0.9332 & 0.9491 \\
pythia-1b              & \textbf{0.9989} & \textbf{0.9974} & \textbf{0.9967} & \textbf{0.9986} 
                       & 0.9510 & \textbf{0.9542} & \textbf{0.9527} & \textbf{0.9531} \\
bloomz-560m            & 0.9302 & 0.9129 & 0.9078 & 0.9363 & 0.8901 & 0.8732 & 0.8675 & 0.9014 \\
bloomz-1b1             & 0.9401 & 0.9244 & 0.9196 & 0.9391 & 0.8993 & 0.8821 & 0.8773 & 0.9038 \\
OPT-350M               & 0.9764 & 0.9717 & 0.9690 & 0.9777 & 0.9437 & 0.9361 & 0.9317 & 0.9440 \\
OPT-1.3b               & 0.9849 & 0.9810 & 0.9774 & 0.9864 & \textbf{0.9525} & 0.9460 & 0.9416 & 0.9552 \\
\bottomrule
\end{tabular}
\end{table*}



\subsection{Federated Learning Framework for Distributed Trajectory Prediction}
\label{sec:federated_learning}

Scalability and data privacy are two critical challenges in deploying real-time IoV systems. Centralized learning architectures often introduce communication bottlenecks and privacy risks, especially when dealing with sensitive, high-frequency vehicular data. FPoTT adopts a federated learning framework to mitigate these limitations, which enables distributed model training across edge devices and a central cloud server without sharing raw data. The architecture comprises two collaborative layers: edge and cloud. The edge layer consists of lightweight, resource-optimized LLMs deployed on vehicles or roadside units. These local models perform real-time inference using localized traffic data, ensuring low latency and immediate responsiveness. The cloud layer, by contrast, maintains a larger, more comprehensive model that aggregates updates from multiple edge clients. It performs global coordination, refining the shared model based on aggregated parameters and disseminating improvements back to the edge. This dual-layer setup enables continuous learning while preserving privacy, as only model weights, not raw data, are transmitted. It also supports dynamic adaptation to local traffic conditions while maintaining consistency at the global level. The formalization of this architecture is presented in Fig.~\ref{fig:fed}, and the parameter update process is described in Equations (3) and (4), as follows:


Let \(\Theta_c\) represent cloud-server model parameters, and \(\Theta_e^{(i)}\) represent parameters for edge devices \(i\). The global model update at the server aggregates local models:
\begin{equation}
    \Theta_c \leftarrow \frac{1}{N}\sum_{i=1}^{N} \Theta_e^{(i)},
\end{equation}
where \(N\) is the number of participating edge devices. Each edge device periodically synchronizes with the global model by:
\begin{equation}
    \Theta_e^{(i)} \leftarrow \alpha \Theta_c + (1-\alpha) \Theta_e^{(i)},
\end{equation}
with \(\alpha\) as the synchronization weight. The illustration of this framework is shown in Fig.~\ref{fig:fed}. Regarding privacy, our federated architecture ensures that raw data remains localized on edge devices. However, recognizing potential threats of gradient leakage and model inversion attacks is crucial. To mitigate these, future extensions of FPoTT may incorporate differential privacy or secure aggregation techniques to further harden the system against adversarial inference.

\subsection{Synthetic NGSIM Data Generator for Traffic Scenario Augmentation}


Real-world vehicular datasets often lack coverage of rare or complex traffic scenarios, which limits the generalization ability of predictive models in dynamic Iot environments. This challenge is tackled through a Transformer-based synthetic data generator modelled after the NGSIM format~\cite{pazho2023vt}, designed to augment training data with diverse and high-fidelity traffic trajectories. The generator improves model robustness and adaptability by simulating underrepresented behaviours such as sudden lane changes, abnormal stops, or traffic surges. The generation process consists of three main stages:


\textbf{Initialization:}
Given initial conditions derived from real traffic data, such as initial positions, velocities, and accelerations, the generator defines an initial state vector:
\begin{equation}
    X_{init} = \{x_1, x_2, \dots, x_m\}, \quad x_i \in \mathbb{R}^{d},
\end{equation}
where \(m\) is the number of vehicles and \(d\) is the dimension of each vehicle state (e.g., position, velocity, acceleration).\\
\textbf{Iterative Prediction:}
The Transformer-based generator iteratively predicts the next state \(\hat{x}_{t+1}\) based on historical states \(\hat{x}_{t}, \hat{x}_{t-1}, \dots, \hat{x}_{t-k}\), as follows:
\begin{equation}
    \hat{x}_{t+1} = \text{TransformerEncoder}(\hat{x}_t, \hat{x}_{t-1}, \dots, \hat{x}_{t-k}; \phi),
\end{equation}
where \(\phi\) denotes the parameters of the Transformer encoder, and \(k\) is the context length defining how many past states are used for prediction. The Transformer encoder captures temporal dependencies and interactions between multiple vehicles, allowing for accurate modelling of complex vehicular dynamics and interactions such as lane changes, accelerations, and decelerations. \\
\textbf{Post-processing and Validation:}
Generated trajectories are further post-processed to ensure realism and compliance with known physical constraints and road regulations. The post-processing step includes trajectory smoothing, collision checking, and velocity normalization:
\begin{equation}
    \hat{X}_{final} = \text{PostProcessing}(\hat{x}_{1:T}),
\end{equation}
where \(T\) is the total length of generated sequences.


\subsection{Training and Inference Pipeline for Real-Time Trajectory Prediction}
\label{sec:training_inference}

FPoTT integrates a tightly coupled training and inference pipeline to ensure accurate and efficient real-time trajectory prediction. This design enables continual adaptation of prompts and model weights based on real and synthetic traffic data, while guaranteeing low-latency performance at the edge. During training, the framework jointly optimizes the Prompt Generator and Core LLM: the former refines prompts using feedback from evaluation metrics, while the latter learns from sequential traffic data sourced from NGSIM and synthetic datasets. At inference time, the optimized prompts guide the Core LLM to deliver accurate, context-aware predictions, enabling responsive and privacy-preserving decision-making on edge devices. The overall framework allows FPoTT to dynamically adapt and optimize vehicle trajectory prediction, ensuring high accuracy, low latency, and robust privacy preservation.

\section{Results and Analysis}
\label{sec:results_analysis}


\subsection{Experiment Setup}
\label{sec:experiment_setup}

\begin{figure}[htbp]
\includegraphics[width=\columnwidth]{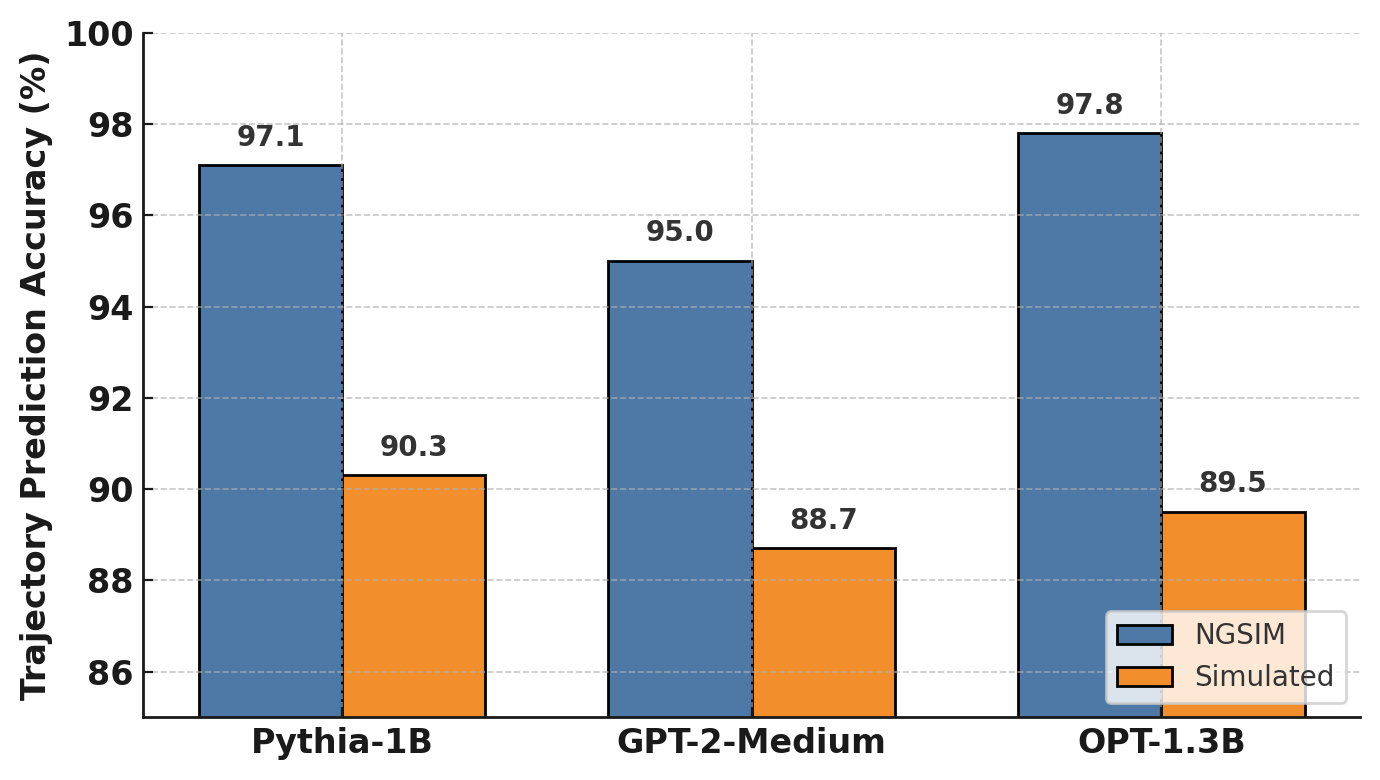}
\caption{Accuracy comparison of different models using default prompts on NGSIM and simulated datasets. EleutherAI/pythia-1b, GPT-2-Medium (355M), and OPT-1.3b are evaluated to assess baseline model performance in IoV trajectory prediction tasks.}
\label{fig:Model Comparison}
\end{figure}

\begin{figure}[htbp]
    \centering
    \includegraphics[width=\columnwidth]{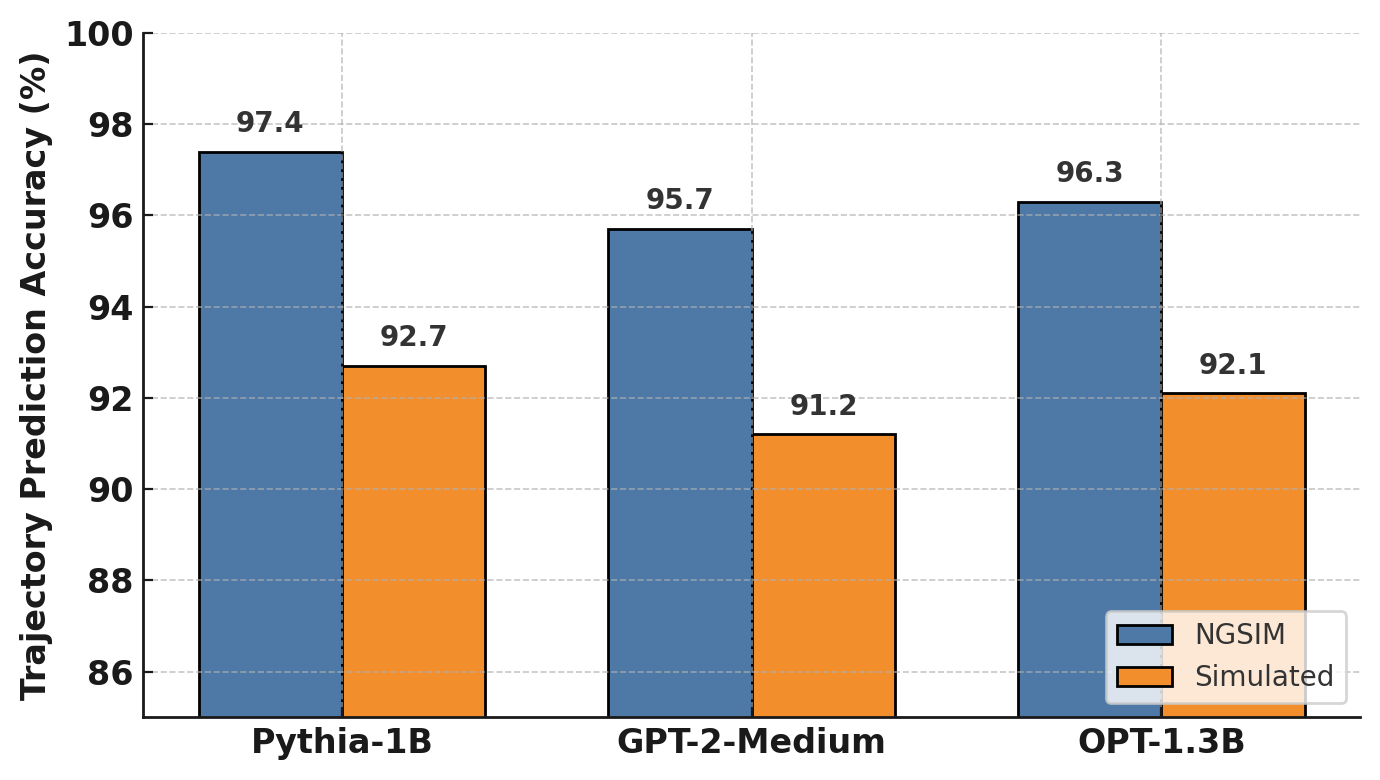}
    \caption{Accuracy comparison of different models under low complexity prompts using NGSIM and simulated datasets. EleutherAI/pythia-1b, GPT-2-Medium (355M), and OPT-1.3b are evaluated to assess model performance in privacy-preserving IoV scenarios.}
    \label{fig:low_complexity}
\end{figure}

\begin{figure}[htbp]
    \centering
    \includegraphics[width=\columnwidth]{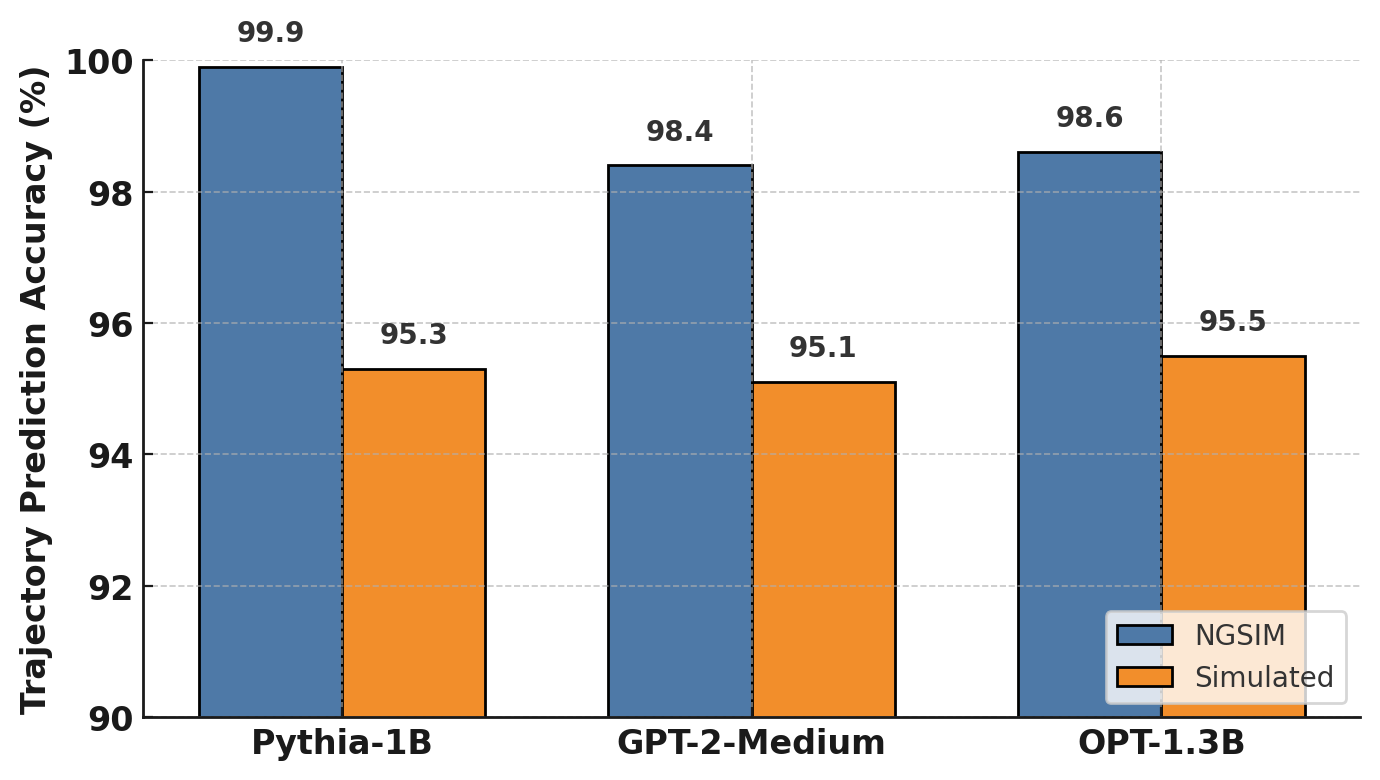}
    \caption{Accuracy comparison of different models under high complexity prompts using NGSIM and simulated datasets. EleutherAI/pythia-1b, GPT-2-Medium (355M), and OPT-1.3b are evaluated to assess the effect of prompt refinement on trajectory prediction in IoV management.}
    \label{fig:high_complexity}
    \vspace{-0.5cm}
\end{figure}

We conducted comprehensive experiments under a high-performance computing environment to validate the proposed FPoTT framework's performance, scalability, and real-time suitability. The goal was to ensure that both training and inference stages could support the demands of dynamic IoV scenarios while maintaining model fidelity. Our setup included Ubuntu 24.04 LTS, an AMD Threadripper 7960X processor, and an NVIDIA A100 GPU (80GB). The system featured 256GB DDR4 RAM and SSD-based storage to support efficient data handling and model training. Network connectivity was provided via a symmetrical 5 Gbps Ethernet connection. The software environment was established using Python 3.10, PyTorch 2.1, and compatible CUDA and cuDNN libraries optimized for maximum GPU acceleration. This configuration ensured efficient utilization of computational resources, allowing the training of complex Transformer-based models and prompt optimization processes. Datasets utilized for training and validation included the publicly available NGSIM dataset and our synthetically generated traffic scenarios. Training processes leveraged parallel data loading and augmentation techniques, maximizing GPU throughput and minimizing training times. Experiments were structured to systematically evaluate different open-source LLMs, including LLaMA, Falcon, and GPT-based architectures such as DeepSeek 7b, OPT, GPT 2, etc., to identify the most suitable models for deployment within our federated learning framework.



\subsection{LLM Selection for Trajectory Prediction}
\label{sec:llm_selection}

To identify the most effective language models for trajectory prediction within the FPoTT framework, we experimented with a range of open-source LLMs. We evaluated their performance across both real-world (NGSIM) and simulated datasets. Table~\ref{tab:performance_comparison_combined} summarizes the classification results, reporting precision, recall, F1-score, and accuracy for each model configuration. The EleutherAI Pythia-1B model demonstrates superior performance on both datasets, achieving the highest accuracy of 99.86\% on NGSIM and 99.99\% on our dataset. Moreover, the smaller variant, Pythia-410M, outperformed other comparable-sized models. Models such as GPT-2-Medium and OPT-1.3b also exhibited strong performance; however, they are still incompatible with Pythia-1B, and similar trends can be observed on smaller models. Conversely, models like Google/flan-t5 exhibited relatively lower metrics, suggesting limited suitability for precise real-time trajectory predictions within IoV contexts. As utilizing the same category of LLMs within a single engineering flow may hinder model diversity, we consider it more reliable to choose an alternative family of LLMs for prompting purposes. We are using GPT-2-Medium as "Prompt LLM" as shown in Fig.~\ref{fig:arch} as it only uses 1/3 of the params compared to OPT-1.3b, and its bigger version GPT-2-Large (774M) does not provide very significant improvements with nearly twice the params included.

\subsection{Prompt Complexity Impact on Model Performance}

To assess the impact of varying prompt complexities generated by our Prompt Generator, we designed a specific set of experiments using two levels of prompt complexity: Low (basic prompts) and High (highly optimized prompts generated by our proposed Prompt Generator), and also compared with the default prompt. We evaluated the EleutherAI/pythia-1b, GPT-2-Medium (355M), and OPT-1.3b models, as these showed promising results in the previous section. Fig.~\ref{fig:Model Comparison} presents a baseline comparison of model performance using default prompts across both the NGSIM and simulated datasets. Among the evaluated models, EleutherAI/pythia-1b consistently demonstrates superior accuracy, exceeding 97\% on NGSIM and over 90\% on the simulated dataset. GPT-2-Medium (355M) and OPT-1.3b also perform well but show more variability between datasets, particularly in their ability to generalize to simulated scenarios. These results establish pythia-1b as a strong candidate for baseline IoV trajectory prediction, even without prompt refinement. Fig.~\ref{fig:low_complexity} examines the effect of low-complexity prompts. The results indicate modest improvements in model performance across all configurations compared to the default prompt setup. Notably, pythia-1b maintains its lead with a slight accuracy gain on both datasets, while GPT-2-Medium and OPT-1.3b show smaller but consistent improvements. This suggests that even lightly structured prompts help clarify input patterns and improve predictive precision, although the overall gains are not yet optimal. Fig.~\ref{fig:high_complexity} captures the impact of high-complexity prompts generated through the proposed Prompt Generator. Here, the models demonstrate significant performance boosts, particularly on the NGSIM dataset. Pythia-1b achieves near-perfect accuracy (above 99.8\%), with comparable gains observed in the simulated dataset. GPT-2-Medium and OPT-1.3b also benefit substantially, validating the effectiveness of context-rich, semantically optimized prompts in enhancing model understanding. These findings underscore the strength of the proposed prompt optimization strategy in boosting LLM-based predictions for IoV applications, especially in privacy-sensitive and real-time environments.



\section{Conclusion}
\label{sec:conclusion}

FPoTT exemplifies the integration of prompt optimization and federated collaboration to enable privacy-preserving, scalable IoV management using open-source LLMs. The framework leverages a prompt optimization mechanism to enhance the effectiveness of trajectory predictions, demonstrating that the quality of prompts plays a critical role in maximizing LLM performance in dynamic vehicular environments. By integrating cloud and edge intelligence through a dual-layer architecture, FPoTT enables real-time decision-making without compromising scalability or privacy. In addition, the framework incorporates a Transformer-based synthetic data generator that enriches the training process with diverse, high-fidelity traffic scenarios, improving model robustness under varied conditions. Comprehensive experiments on real and simulated datasets highlight the framework’s accuracy, adaptability, and practical viability, with the EleutherAI Pythia-1B demonstrating state-of-the-art performance exceeding 99.8\% accuracy. These results affirm the potential of LLM-driven federated intelligence to transform IoV systems, offering a secure, cost-effective, and future-ready solution for intelligent transportation.




\end{document}